\documentclass{article}

\PassOptionsToPackage{numbers,compress}{natbib}

\usepackage[final]{neurips_2019}

\usepackage[utf8]{inputenc} % allow utf-8 input
\usepackage[T1]{fontenc}    % use 8-bit T1 fonts
\usepackage{hyperref}       % hyperlinks
\usepackage{url}            % simple URL typesetting
\usepackage{booktabs}       % professional-quality tables
\usepackage{amsfonts}       % blackboard math symbols
\usepackage{nicefrac}       % compact symbols for 1/2, etc.
\usepackage{microtype}      % microtypography

\usepackage{amsmath}
\usepackage{amssymb}
\usepackage{graphicx}
\usepackage{todonotes}
\usepackage{caption}
\usepackage{subcaption}

\title{bLIMEy:\\Surrogate Prediction Explanations Beyond LIME}

\author{%
  Kacper Sokol\\
  Department of Computer Science\\
  University of Bristol\\
  Bristol, United Kingdom\\
  \texttt{K.Sokol@bristol.ac.uk}\\
  \And
  Alexander Hepburn\\
  Department of Engineering Mathematics\\
  University of Bristol\\
  Bristol, United Kingdom\\
  \texttt{ah13558@bristol.ac.uk}\\
  \AND
  Raul Santos-Rodriguez\\
  Department of Engineering Mathematics\\
  University of Bristol\\
  Bristol, United Kingdom\\
  \texttt{enrsr@bristol.ac.uk}\\
  \And
  Peter Flach\\
  Department of Computer Science\\
  University of Bristol\\
  Bristol, United Kingdom\\
  \texttt{Peter.Flach@bristol.ac.uk}
}

\begin{document}

\maketitle

\begin{abstract}
Surrogate explainers of black-box machine learning predictions are of paramount importance in the field of eXplainable Artificial Intelligence since they can be applied to any type of data (images, text and tabular), are model-agnostic and are post-hoc (i.e., can be retrofitted). %
%LIME and surrogate explanations are not the same!
The Local Interpretable Model-agnostic Explanations (LIME) algorithm is often mistakenly unified with a more general framework of surrogate explainers, which may lead to a belief that it is the solution to surrogate explainability. %should not be taken at their face value
In this paper we empower the community to ``build LIME yourself'' (bLIMEy) by proposing a principled algorithmic framework for building custom local surrogate explainers of black-box model predictions, including LIME itself. %
To this end, we demonstrate how to decompose the surrogate explainers family into algorithmically independent and interoperable modules and discuss the influence of these component choices on the functional capabilities of the resulting explainer, using the example of LIME.%
%Our study is supported by experiments done using a modular open source implementation of bLIMEy in Python, which publish alongside this manuscript.%
\end{abstract}

\section{Introduction}
% Here's a problem
Local Interpretable Model-agnostic Explanations (LIME) \cite{ribeiro2016why} is a popular technique for explaining predictions of black-box machine learning models. % predictive
It greatly improves on \emph{surrogate explanations} \cite{craven1996extracting} by introducing \emph{interpretable data representations}, hence making it applicable to image and text data in addition to tabular data. %
Images can be represented as a collection of super-pixels and translated into a binary on/off vector indicating which super-pixel stays the same and which is occluded (removed). %
Equivalently, text can be represented as a bag of words translated into a similar on/off binary vector. %
Furthermore, LIME can be used as a fast approximation of SHapley Additive exPlanations (SHAP) \cite{lundberg2017unified} as the latter is computationally expensive -- the cost of providing various guarantees for the produced explanations. %
However, the adoption of LIME is limited since it is provided as a monolithic explainability tool with little room for customisation when, in reality, it is just one possible realisation of the highly modular \emph{surrogate explanations} framework. % (discussed in Section~\ref{sec:modularity}). %
We argue that allowing the user to make informed choices and build a custom surrogate explainer that is designed for a specific task can greatly improve the quality of produced explanations, therefore warrant a wider adoption of surrogate explanations.%

% It's an interesting problem
Since surrogate explanations are model-agnostic, they can be applied to any predictive system, hence have a high impact in the field of eXplainable Artificial Intelligence if they become accessible, accountable and accurate. %
According to the ``no free lunch'' theorem, a single solution can never perform better than all the other approaches across the board. %
However, by allowing the user to take advantage of surrogate explanations' modularity, therefore customising them for a problem at hand, we can achieve the best possible explanation that the surrogate explainers family can offer. %
We can further improve the quality of the explanations by educating the users on the possible component choices, their properties, influence on the explanations, advantages and caveats.%

% It's an unsolved problem
To the best of our knowledge, LIME is the only available surrogate explainability tool and modifying its default behaviour often requires tinkering with LIME's source code what may discourage some of the practitioners. %
% Here's our idea
We address these shortcomings by taking advantage of surrogate explanations' modularity to create a unified algorithmic framework for building this type of explainers, which we call \textbf{bLIMEy} -- \textbf{b}uild \textbf{LIME} \textbf{y}ourself. %
A range of possible algorithms can be used for each module -- with the choices discussed in Section~\ref{sec:discussion} -- creating a suite of customisable surrogate explainers. %
Their varying capabilities and restrictions greatly influence the resulting surrogate explainer, therefore each of them should be accompanied by a critical discussion and usage suggestions. %
% How does the idea work
To this end, we have decomposed the surrogate explanation framework into independent algorithmic components. %
We implemented a choice of algorithms for each of them in Python under the BSD 3-Clause open source licence, therefore allowing for their commercial use. %
Our implementation is accompanied by a ``how-to'' guide\footnote{\url{https://fat-forensics.org/how_to/transparency/tabular-surrogates.html}} %
outlining how to compose custom surrogate explainers and discussing pros and cons of selected component choices. %
It is also capable of recreating the LIME algorithm for tabular, image and text data in a way that mitigates most of its issues reported in the literature \cite{fen2019should,lakkaraju2019faithful,laugel2018defining}.%

% How it compares to other people's ideas
%Existing research in this direction includes investigating instability and randomness of LIME explanations \cite{fen2019should,lakkaraju2019faithful}, however it does not pinpoint its root cause. %
%Other researchers have proposed minor modifications of the LIME algorithm \cite{laugel2018defining}, which they show to have positive effects on the quality of the resulting explanations, however their alterations unintentionally compromise the integrity of LIME making the two methods incomparable and the improvements not applicable to more general cases beyond the specific ones presented in their research.%
%
%\citet{laugel2018defining} have tried to improve on it by altering its data sampling procedure, however their approach compromised the locality of LIME samples by disabling the discretisation step, which is responsible for generating interpretable representation of the data (see below for more details). %%Section~\ref{sec:related}). %
%
Our research on surrogate explanations \cite{craven1996extracting} was inspired by manuscripts investigating the instability and sources of randomness \cite{fen2019should,lakkaraju2019faithful} in LIME explanations\footnote{Produced with LIME's official open source implementation: \url{https://github.com/marcotcr/lime}.} \cite{ribeiro2016why}, which could not pinpoint the root cause of this undesired and detrimental behaviour. %
\citet{laugel2018defining} attempted to ``fix'' LIME for tabular data by replacing its sampling method with an explicitly local sampler, however their experiments used LIME with disabled discretisation (responsible for generating interpretable data representation), therefore unintentionally compromising the integrity of the algorithm making the two methods incomparable and the improvements not applicable to more general cases beyond the specific ones presented in their research. %
%which could only use proximity scores (kernelised distances) to enforce locality of the explanation.% -- this is not true the normal sample is local -- around the data point
\citet{henin2019towards} introduced a unified (theoretical) framework that allows for systematic comparison of black-box explainers by characterising them alongside two dimensions: \emph{data sampling} and \emph{explanation generation}. %
In contrast, our (practical) approach is focused on algorithmic and implementation aspects of the \emph{surrogate} subset of the black-box explainers family and extends \citeauthor{henin2019towards}'s decomposition with a third dimension -- \emph{interpretable representation} -- therefore bridging the gap between LIME and surrogate explanations \cite{craven1996extracting}.% the original

\section{bLIMEy: Modular Surrogate Explanations\label{sec:modularity}}
% LIME overview
Before delving into the bLIMEy framework we encourage the reader to consult Appendix~A1 for an overview of the LIME explainer architecture. %
%
%\subsection{bLIMEy}
The bLIMEy framework decomposes surrogate explanations of a black-box model prediction for a selected data point into three distinct steps:%
%the following three steps: interpretable data representation, data sampling and explanation generation.%
\begin{description}
    \item[Interpretable Data Representation] %
%\subsubsection{Interpretable Data Representation}
Transformation (possibly bidirectional) from the original data domain (i.e., feature space) into an interpretable domain (and back). %
This step is optional for tabular data but required for image and text data. %
Interpretable domains tend to be \emph{binary vectors} encoding presence or absence of human-comprehensible characteristic in the data.% observed indicating original domain

    \item[Data Sampling] %
%\subsubsection{Data Sampling}
Data augmentation (sampling) in the neighbourhood of the data point selected to be explained. %
For images and text data sampling must be performed in the interpretable domain while for tabular data it can be done in either of the domains. %
Next, the sampled data must be predicted with the black-box model. %
If the data were sampled in the interpretable domain, they need to be \emph{reverted} back to the original representation to complete this task.%

    \item[Explanation Generation] %
%\subsubsection{Explanation Generation}
An inherently interpretable model is trained on the locally sampled data, which is used to explain the selected data point. %
If an interpretable (binary) representation is used, XNOR\footnote{\(1\) if the same and \(0\) if different.} can be applied between the selected data point and the sampled data to focus the local model on presence and absence of these interpretable characteristics. %
The interpretable features for which the value of the selected data point is \(0\) can be safely removed to reduce the dimensionality. %
This task enforces the locality of the sampled data and introduces sparsity to the explanation\footnote{The XNOR operation and dropping \(0\)-valued interpretable features do not affect image and text data as the selected data point is always represented as an all-\(1\) binary vector in the interpretable domain.}. %
Sparsity can be enforced even further by applying a dimensionality reduction technique. % (to either representation). %
To further enforce the locality of the explanation, when training the local model the sampled data can be weighted based on a kernelised distance between the chosen data point and the sampled data (in either representation).%
\end{description}

\section{Discussion\label{sec:discussion}}
When building a surrogate explainer every module choice may limit its overall functionality and the range of algorithms supported by other modules. %
Here, we discuss (often unintended) consequences of choosing a particular algorithm for each module and exemplify how such a choice has affected the explainer using the example of LIME. %In this sub-section  architecture
%
%
%LIME is commonly misinterpreted.
%
%\subsection{Surrogate Tree Explainer for Tabular Data -- LIME Alternative}
%Downsides of lime and how we can improve on that.
%the iris data set the two moons data set
We support our discussion with an empirical comparison of various data samplers (Appendix~A2) and an illustration of how a tree-based surrogate improves over a linear one, i.e., LIME, for tabular data without an interpretable representation (Appendix~A3).%algorithms sampling methods decision 
%We support all of these guidelines for building surrogate explainers by comparing the behaviour of various sampling algorithms for the Iris data set \cite{fisher1936use} in Appendix~A2. %(tabular)  compairson of sampling methods for the Iris data set.
%Appendix~A3 shows a decision tree-based surrogate as an improvement over a linear model, i.e., LIME, using the two moons data set -- a safe alternative for tabular data without interpretable representation (discretisation).%

% Interpretable representation
To avoid unnecessary randomness affecting the explanations produced by surrogate explainers, the data transformation from their original domain \(\mathcal{X}\) into an interpretable representation \(\mathcal{X}^\prime\) must be \textbf{bijective} -- the mapping from \(\mathcal{X}\) to \(\mathcal{X}^\prime\) has to be a \emph{one-to-one correspondence} -- %\emph{unique}
and it must have a corresponding and uniquely defined \textbf{inverse function} -- a data point in \(\mathcal{X}^\prime\) can be translated into a unique data point in \(\mathcal{X}\). %
% bidirectionality \emph{reversible} \emph{bidirectional}
In LIME, the interpretable representation for both image and text data satisfies these two requirements. %
A sentence can be easily represented as a binary vector indicating presence or absence of unique words in that sentence and such a binary vector can be transformed into (bag-of-words representation of) a sentence. %
Similar reasoning applies to images where a binary vector indicates whether a super-pixel (large, non-overlapping chunks of an image) in an image should have the same pixel values as the original image or be occluded with an arbitrary patch or a solid colour. %
For tabular data, an interpretable representation is achieved by discretisation and binarisation (one-hot encoding), in which case bijection is preserved but the inverse function is ill-defined. %
%sampling from interpretable domain of tabular data may have unintended consequences.
While the binarisation of categorical features (there is no need for discretisation) is invertible, numerical features that have been discretised by binning (and binarised) cannot be uniquely reconstructed into their original representation \(\mathcal{X}\). %
LIME resolves this by sampling from a normal distribution (with clipping at bin boundaries) fitted to each numerical bin for each sampled data point to reconstruct it in the original domain (\(\mathcal{X}\)) -- the unidentified source of randomness reported by \citet{fen2019should}.%
%\todo{Alex: Please verify this statement. FINE}
%difficult ot compare as different domains unles the data points are perfectly symetric and recoverable

This undesired behaviour for tabular data is a consequence of transforming the data into their interpretable representation \(\mathcal{X}^\prime\) first and then sampling. %
While such order is required for image and text data -- it would be meaningless to sample from a grid of pixels or a sequence of characters respectively -- it is not compulsory for tabular data, therefore providing an opportunity to avoid the ``reverse sampling'' step. %
By sampling first (in \(\mathcal{X}\)) and transforming the tabular data into an interpretable representation later (\(\mathcal{X}^\prime\)), the inversion of the latter step is no longer required since both of the representations are available. %
This order of operations, however, requires the sampling procedure to be ``as much local as possible'' since sampling from the interpretable domain implicitly introduces the locality\footnote{For text data this is sampling from within the same sentence by leaving the words intact or removing them and for images this is modifying the original image by occluding its parts. For tabular data LIME achieves the sample locality in this step by applying the XNOR operation explained in the next paragraph.}. %
While any sort of sampling should suffice in the interpretable domain (as long as the XNOR filtering is performed -- see the next paragraph) a local sampling method, e.g., MixUp \cite{zhang2018mixup} or Growing Spheres \cite{laugel2018spheres}, is required when it is performed in the original data domain (for tabular data) -- see the results shown in Appendix~A2. %
Furthermore, sampling in this domain requires the sampling method to produce data points that are assigned more than one class (or significantly different class probabilities for probabilistic models) %
%a different class than the one of the explained data point %(e.g., significantly different class probabilities)
by the black-box model, otherwise a \emph{meaningful} local surrogate model cannot be fitted. %
%%%
Given the random nature of the sampling procedure, the only way to ensure reproducible explanations is to always have the same local sample, which can only be achieved by fixing the random seed.% prior to this step

While reducing the dimensionality of the interpretable domain for image and text data is detrimental for the explanation -- e.g., ``black holes'' in images and missing words in sentences -- it is recommended for tabular data to reduce the number of ``important factors'' presented to the explainee. %
When an interpretable representation is not used for tabular data, dimensionality reduction should be considered a necessity. %
In LIME, sparsity (and locality) of an explanation for tabular data is partially achieved -- when using an interpretable representation -- by dropping all of the interpretable features for which the feature value of the explained data point in the interpretable domain is \(0\). %
This operation is equivalent to keeping only the categorical feature values and numerical feature bins in which the explained data point resides (the aforementioned implicit locality for tabular data). %lies
%\todo{Alex: Is this correct? In particular for categorical features. - only if discretisation is used}
For example, for two numerical features \((x_1, x_2)\) and their interpretable representations \((x_1 < 2, 2 \leq x_1 < 7, 7 \leq x_1)\) and \((x_2 < 0, 0 \leq x_2)\) and an explained data point \((4, 2)\) -- \((0, 1, 0, 0, 1)\) in the interpretable representation -- only \(2 \leq x_1 < 7\) and \(0 \leq x_2\) dimensions would be preserved. %
%for LIME and interpretabel representation of tabular data reducing dimensions means picking the range of features
This step combined with transforming the interpretable feature space by applying the XNOR operation between the explained data point and the sampled data is \textbf{required} for proper functioning of LIME since its goal is to explain a prediction by quantifying the (positive or negative) effect of changing any of the interpretable feature ranges within which the explained data point resides, hence the choice of a \emph{linear model} for the local surrogate and the use of its coefficients as the explanation. %
%LIME explains predictions of data points in terms of importance of characteristics present in this data point.
Therefore, the question that the LIME explanation tries to answer is whether for the current black-box classification of the selected data point the value of \(x_1\) between \(2\) and \(7\) has a positive or a negative effect when compared to \(x_1\) being outside of this range. %
%For example, for the current classification of a selected point, does \(x_1\) between \(2\) and \(7\) have a positive or a negative effect when compared to \(x_1\) being outside of this range. %
Similarly, for a particular classification of an image or a sentence, does a given super-pixel or a word have a positive or a negative effect, i.e., would removing this super-pixel or word change the black-box classification outcome. %
Thereby using LIME for tabular data without an interpretable representation forfeits the locality of the sample introduced by applying the XNOR feature transformation in which case the explainer relies purely on kernelised distance weighting and normal data sampling around the explained data point (albeit with sampling variance for each feature calculated based on the whole data set, which decreases the locality effect) to induce the explanation locality. %
%\todo{Alex: could you please validate this? - true}
An interpretable representation for tabular data may be skipped altogether in which case other modules of the surrogate explainer, like the data sampler (cf.\ Appendix~A2), should guarantee the locality of an explanation.% ... the choice of a surrogate ...

%%% Reviewer 1: Can conclusions be made about which explanations are better than others? Can that be explained so the reader understands what conclusions will be drawn from different explanations and the implications of drawing different conclusions from different explanations?
%Last but not least, the surrogate model can be trained either as a regressor on the probabilities outputted by the black-box model (as in LIME), in which case it has to model (explain) a single class selected by the user (one vs.\ the rest); or a classifier trained on all of the possible classes or, again, one vs.\ the rest.% explainig all of them together not possible by a linear model
The surrogate model can be trained in a number of different ways: as a regressor of the probabilities outputted by the black-box model (as in LIME), in which case it has to model (explain) a single class selected by the user; or as a classifier when the black-box model is a thresholded probabilistic model or a classifier. %
Another choice is the training scheme: the surrogate model can either be trained as a multi-class or one-vs-rest predictor, with the latter approach being required for surrogate regressors of black-box probabilistic predictors. %
The choice of a surrogate model is also important; if local feature importance (or interpretable feature influence) is desired, a linear model is a good pick as long as all of the features are normalised to the same range (LIME satisfies this by using the interpretable binary representation or otherwise explicitly normalising the features) and these features are ``reasonably'' independent. %vanilla LIME satisfies this
A lack of normalisation causes the feature weights to be \emph{incomparable}, therefore rendering the explanation uninformative. %extracted from the surrogate model 
While explainability of linear models is limited to feature importance, a different type of an explanation -- logical conditions outlining the behaviour of a black-box model in the neighbourhood of the selected data point -- can be generated with a surrogate decision tree (cf.\ Appendix~A3). %
The selection here should be motivated by the desired type of the explanation -- e.g., ``Why class A?'' vs ``Why class A and not B?'' -- and its format -- a feature importance bar plot vs a conjunction of logical conditions -- which are problem-dependant.%and the targeted audience. %
%This choice also influences the type of explanation that it can produce: while linear models are only capable of outputting feature importance, decision trees, for example, can additionally produce logical conditions (see Appendix~A3). %
% one class coutnerfactuals or (interpretable) features important for one class
%%% model choice -> explanation type choice: feature importance vs. logical conditions

\section{Conclusions and Future Work}
In this paper we introduced bLIMEy: a modular algorithmic framework for building custom surrogate explainers of black-box predictions. %
We discussed dangers associated with algorithmic choices for each of its modules and showed how to avoid common pitfalls. %provided recommendations to help their creators
bLIMEy is accompanied by an open source implementation that includes a selection of algorithms for every module of the framework, therefore empowering the community to build surrogate explainers customised to the task at hand.%
%giving its users an opportunity to easily build a surrogate explainer suitable for any task.%
%unleashing creative power to the user.%
%
%Test various combinations and inspect them to provide the community with recommendations how to build these surrogates based on the problem at hand, classification, regression, tabular data, images, text, etc.
%By analysing components of surrogate explanations provide recommendations

In the future we will investigate the behaviour of various surrogate models in high-dimensional spaces and design a range of metrics to measure the quality and stability of surrogate explanations. %
We will provide one measure for each of the following three competing objectives: (1) local approximation of the closest decision boundary; (2) the ability to mimic the black-box model locally; and (3) the global faithfulness of the local surrogate model, which will engender trust in the explanations and mitigate the need for user studies, which lack universally agreed objective and often entail confirmation bias.% properties
%1.  -- how well does it approximate the closest global decision boundary
%2. how well does it perform in the vicinity of the explained data point
%3. What the global faithfulness

%\subsubsection*{Acknowledgments}
%Use unnumbered third level headings for the acknowledgments.

\bibliographystyle{plainnat}%plain}
{\small\bibliography{neurips_2019}}

%\clearpage
\newpage
\section*{Appendix}
All of the results presented in this appendix can be reproduced with a Jupyter Notebook hosted in the bLIMEy's GitHub repository\footnote{\url{https://github.com/So-Cool/bLIMEy/tree/master/HCML_2019}}. %
This notebook can be executed online using Binder\footnote{\url{https://mybinder.org}} by following the URL placed in its top cell. %
The experiments were done using FAT Forensics \cite{sokol2019fat} -- an open source package implementing various fairness, accountability and transparency algorithms. %
The description of each function used for these experiments can be found in the API\footnote{Application Programming Interface} documentation of FAT Forensics\footnote{\url{https://fat-forensics.org/api.html}}.%

\subsection*{A1: LIME Overview}
% LIME overview
%inspect LIME components and their interconnections
%LIME, the dominant realisation surrogate explanations for predictions of black box models behaves as follows...
Before discussing the LIME algorithm we examine the concept of LIME's interpretable data representations. %
For text data the interpretable representation is achieved by encoding a sentence as a bag of words and representing it as a binary vector, which indicates presence and absence of unique words. %
Such interpretable word vectors can then be represented as sentences by removing words for which the value in this binary vector has been changed to \(0\). %
The interpretable representation for images is generated by dividing images into super-pixels -- non-overlapping segments -- each one encompassing a part of the image that represents a concept (e.g., an object) meaningful to humans. % patches of the image
Such interpretable image vectors can be then represented as images by occluding the super-pixels for which the value in this binary vector has been set to \(0\). %
Finally, tabular data can be transformed into the interpretable representation by one-hot encoding the categorical features and binning the numerical ones. % 
Such interpretable tabular data vectors can be then represented in the original feature domain by altering the feature values accordingly to the changes made in the binary vector. %
This important concept of interpretable data representations enables LIME to explain data (i.e., raw features such as pixel values for images) or their representations (internally used by black-box predictive models, such as high-dimensional word embeddings) that are inherently human-incomprehensible.%not understandable

Therefore, for a data point \(x \in \mathcal{X}\) to be explained and an arbitrary black-box probabilistic predictive model \(f: \mathcal{X} \rightarrow \mathcal{Y}\) the default LIME algorithm proceeds as follows:%
\begin{enumerate}
    \item Find the human interpretable representation \(x^\prime \in \mathcal{X}^\prime\) of the data point \(x\) chosen to be explained, where \(\mathcal{X}^\prime\) denotes the interpretable domain.%
    \item Sample data from the interpretable domain \(\mathcal{X}^\prime\). For image and text data this is done by uniformly replacing \(1\)'s in \(x^\prime\) with values from \(\{0, 1\}\) set to get new data points in the ``neighbourhood'' of \(x\), e.g., by randomly occluding super-pixels in an image or removing words from a sentence. Tabular data is first discretised into a representation where categorical features are left unchanged and each numerical feature is transformed into a categorical feature that indicates the numerical bin (interpretable representation) to which this feature belongs, e.g, \(\hat{x}_3 = 1\) for \(\{(-\infty, 0.5), [0.5, 1.3), [1.3, \infty)\}\) bins indicates \(x_3 \in [0.5, 1.3)\). This representation (\(\widehat{\mathcal{X}}\)) is used for tabular data sampling to avoid assigning a sample to two different bins for a single feature, what could have happened had the sampling been performed in the binary representation (\(\mathcal{X}^\prime\)) -- where each bin for each feature is represented as a separate binary feature, e.g., \((x_{3-0}^\prime, x_{3-1}^\prime, x_{3-2}^\prime)\) sampled in a binary domain and resulting in a \((0, 1, 1)\) vector would give \(x_3 \in [0.5, 1.3)\) and \(x_3 \in [1.3, \infty)\). After the sampling step the tabular data is transformed into the binary interpretable representation \(\mathcal{X}^\prime\).%in a random fashion
    \item Invert the representation of the sampled data from \(\mathcal{X}^\prime\) to \(\mathcal{X}\) and predict their probability for a selected class \(c\) with the black-box model \(f\). Usually, \(c\) is selected to be the class assigned to the explained data point \(x\) by the black-box model \(f\).%
    \item Drop all of the interpretable features for which the value of the explained data point (in the interpretable domain) is \(0\), therefore creating a new representation \(\widetilde{\mathcal{X}}\) -- this introduces sparsity of the explanation and enforces its locality (see Section~\ref{sec:discussion} for more details).%
    \item Calculate the distances between the sampled data and the explained data point in \(\widetilde{\mathcal{X}}\) and kernelise them (using the exponential kernel) to serve as proximity/similarity scores used to weight the sampled data during the training of the local surrogate model (to enforce locality of the explanation).% %%% the distances for the kernel are in the binarised space
    \item Compute logical XNOR between the new interpretable representation \(\widetilde{\mathcal{X}}\) of the explained instance \(\tilde{x}\) and all of the sampled data points \(\tilde{x}_{\text{sampled}}\) to create a data set \(\widetilde{\mathcal{X}}_\text{XNOR}\) which prescribes the effect of change of a data point in the interpretable domain on its classification outcome (see Section~\ref{sec:discussion} for more details).%
    \item Use K-LASSO to further limit the number of features used in the explanation and train a linear regression on \(\widetilde{\mathcal{X}}_\text{XNOR}\) using the black-box predictions computed before as the target (i.e., probabilities of the previously selected class \(c\)). The coefficients of this model (feature weights) are used to interpret the (positive or negative) importance of each human-comprehensible feature.%
\end{enumerate}

For more details please consult the LIME manuscript \cite{ribeiro2016why} and its official open source implementation\footnote{\url{https://github.com/marcotcr/lime}}.

\subsection*{A2: Comparison of Data Samplers for Tabular Data}
\begin{figure}[htb]
    \centering
    \includegraphics[width=0.45\textwidth]{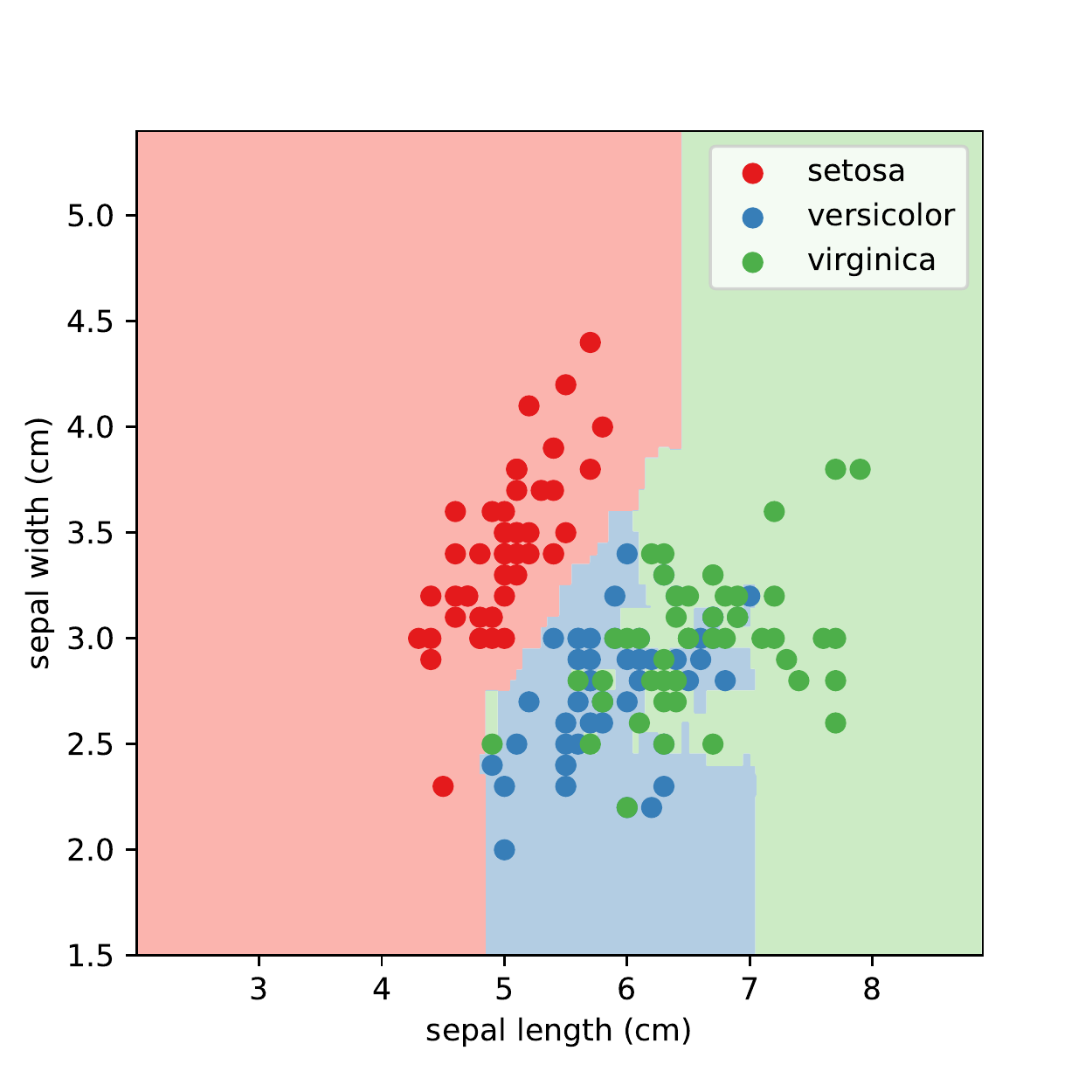}
    \caption{The Iris data set plotted alongside \emph{sepal length (cm)} on the x-axis and \emph{sepal width (cm)} on the y-axis with the markers colour-coded based on the ground truth annotation. The background shading represents the decision boundary of the underlying black-box model (a random forest classifier).\label{fig:iris}}
\end{figure}

To show the behaviour of different data sampling methods we use the Iris data set \cite{fisher1936use}. %
We plot the data alongside two dimensions -- \emph{sepal length (cm)} on the x-axis and \emph{sepal width (cm)} on the y-axis -- to facilitate easy visual comparison. %
The three colours visible in the plots represent the three classes of the Iris data set: setosa, virginica and versicolor. %
The data set plotted alongside these two features with the markers colour-coded based on the ground truth annotation is shown in Figure~\ref{fig:iris}. %
The background shading in this plot indicates the decision boundary of the underlying black-box model (a Random Forest Classifier trained with scikit-learn \cite{scikit-learn}).%

\begin{figure}[htb]
    \centering
    \begin{subfigure}[b]{0.45\textwidth}
        \centering
        \includegraphics[width=\textwidth]{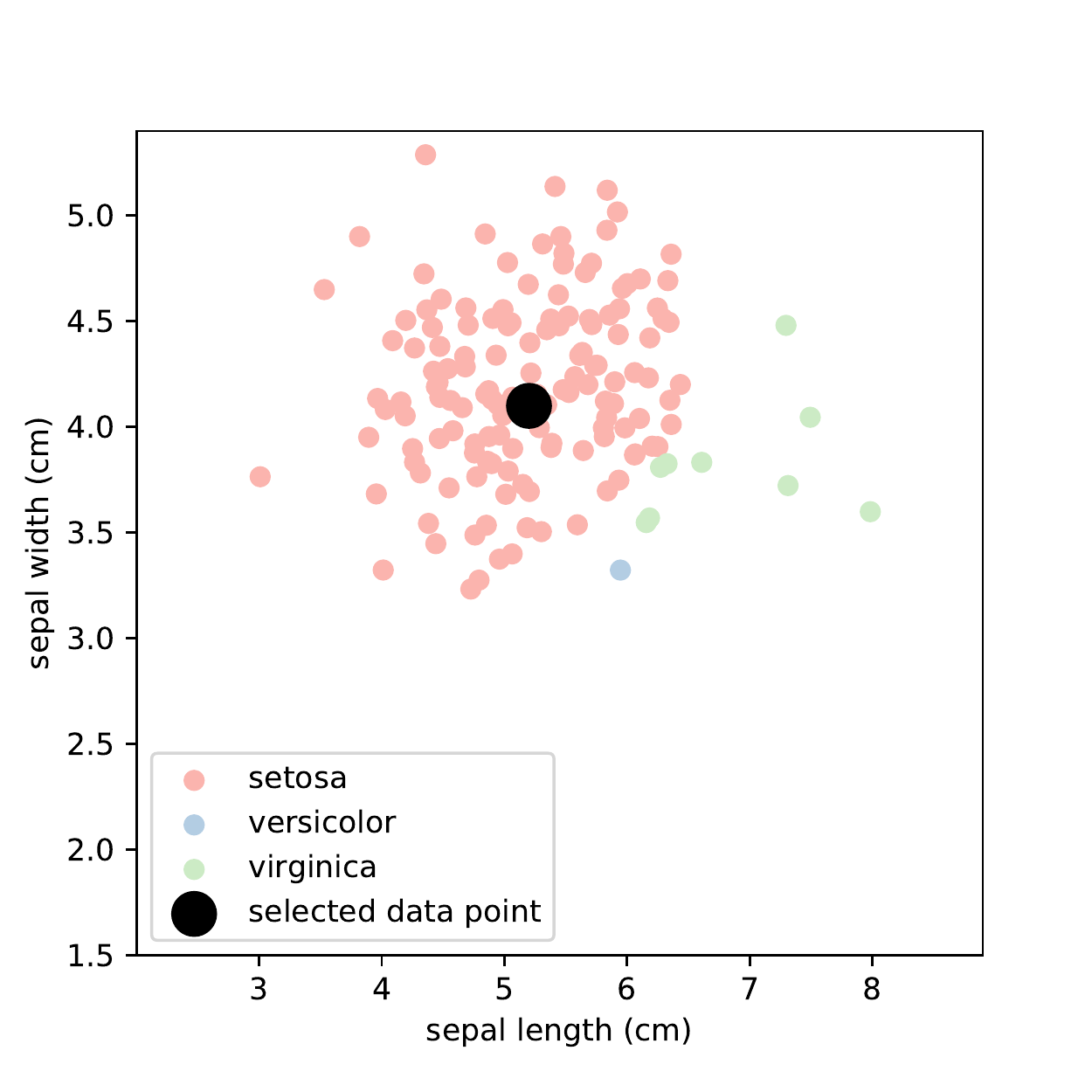}
        \caption{Normal.}
        \label{fig:normal}
    \end{subfigure}
    \hfill
    \begin{subfigure}[b]{0.45\textwidth}
        \centering
        \includegraphics[width=\textwidth]{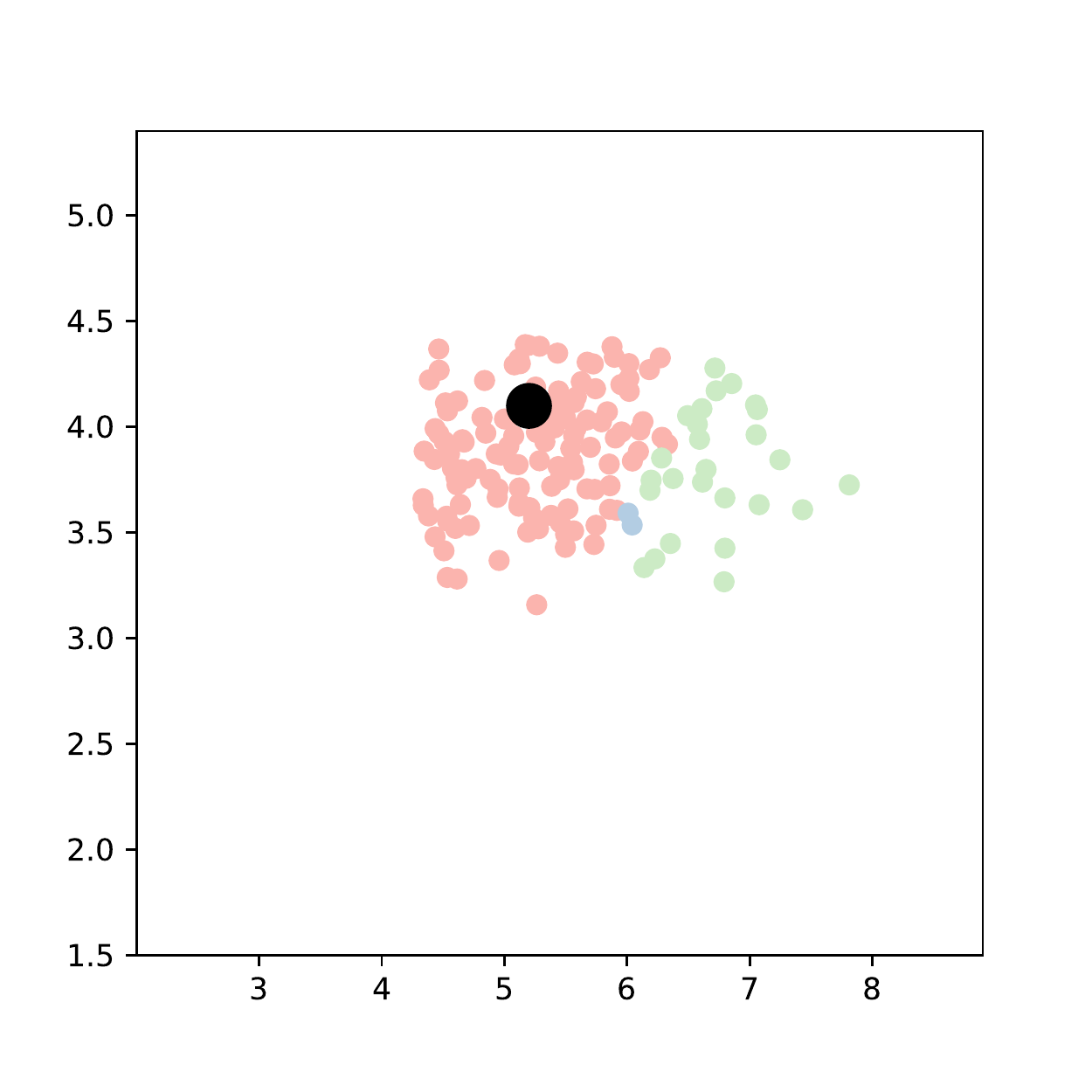}
        \caption{Truncated normal.}
        \label{fig:truncated_normal}
    \end{subfigure}
    \newline
    \begin{subfigure}[b]{0.45\textwidth}
        \centering
        \includegraphics[width=\textwidth]{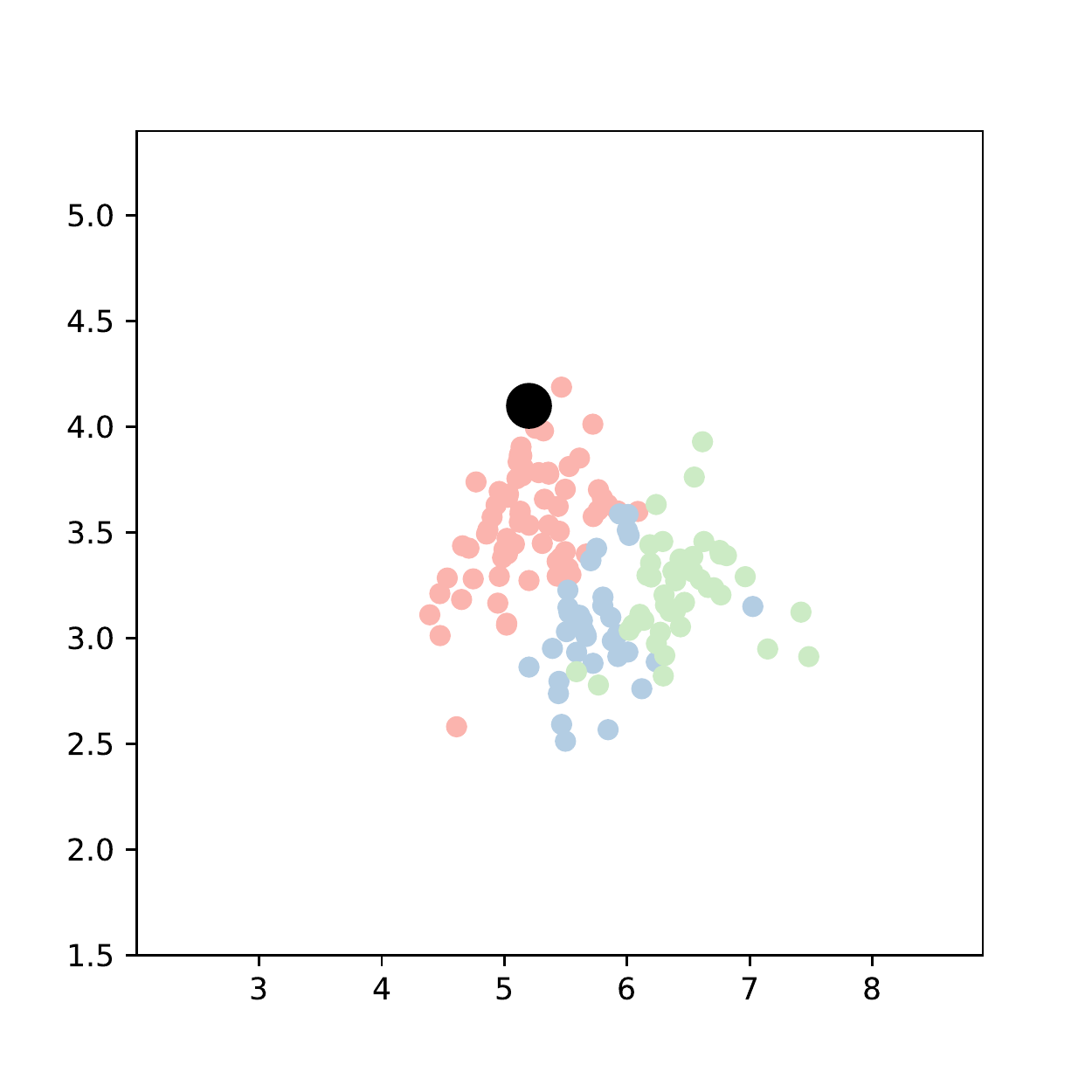}
        \caption{MixUp.}
        \label{fig:mixup}
    \end{subfigure}
    \hfill
    \begin{subfigure}[b]{0.45\textwidth}
        \centering
        \includegraphics[width=\textwidth]{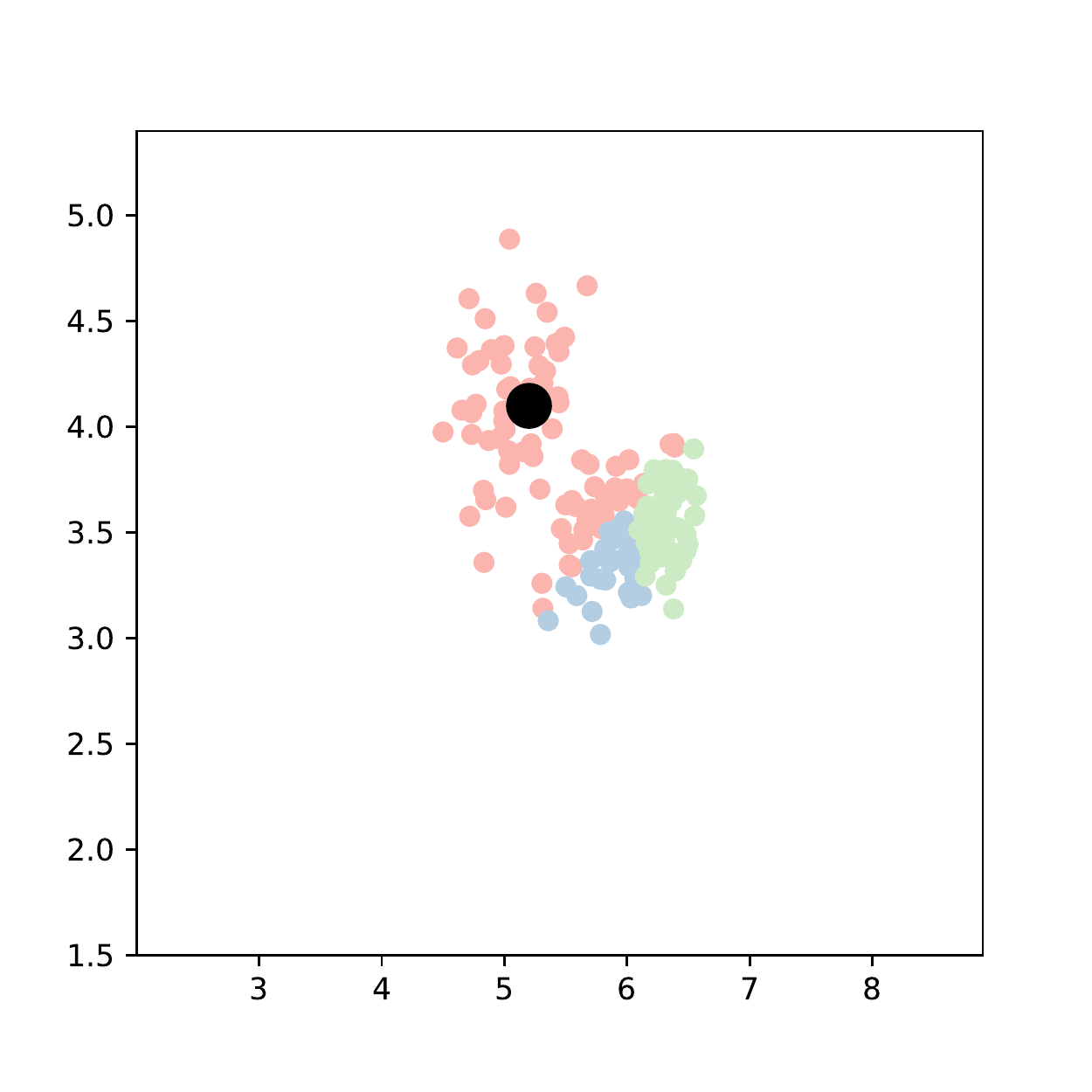}
        \caption{Normal Class Discovery.}
        \label{fig:normal_class_discovery}
    \end{subfigure}
       \caption{The effect of different sampling algorithms on the locality of the sample for the Iris data set plotted along \emph{sepal length (cm)} on the x-axis and \emph{sepal width (cm)} on the y-axis. The black dot is the explained data point for which the sample is generated. Red, blue and green dots are the predictions (the three classes of the Iris data set) assigned by the underlying black-box model (cf.\ Figure~\ref{fig:iris}).}
       \label{fig:sampling_effect}
\end{figure}

We initialise each of the samplers with the full Iris data set and generate 150 data points around the selected instance: the black dot. %
Figure~\ref{fig:sampling_effect} shows the importance of choosing an appropriate sampling method when building a surrogate explainer. %
Some of the data samplers may have difficulties locating the closes decision boundary (see, for example, Figure~\ref{fig:normal} and \ref{fig:truncated_normal}) when the selected data point is far from any decision boundary of the black-box model (which may be common in high-dimensional spaces due to the curse of dimensionality), therefore generating data for which fitting a local surrogate model may not be possible. %
Another important aspect of the sampled data is their clear class imbalance, which needs to be accounted for during the training procedure of the surrogate model.%

\subsection*{A3: Decision Tree-based Surrogate Explainer for Tabular Data}
\begin{figure}[htb]
    \centering
    \includegraphics[width=0.55\textwidth]{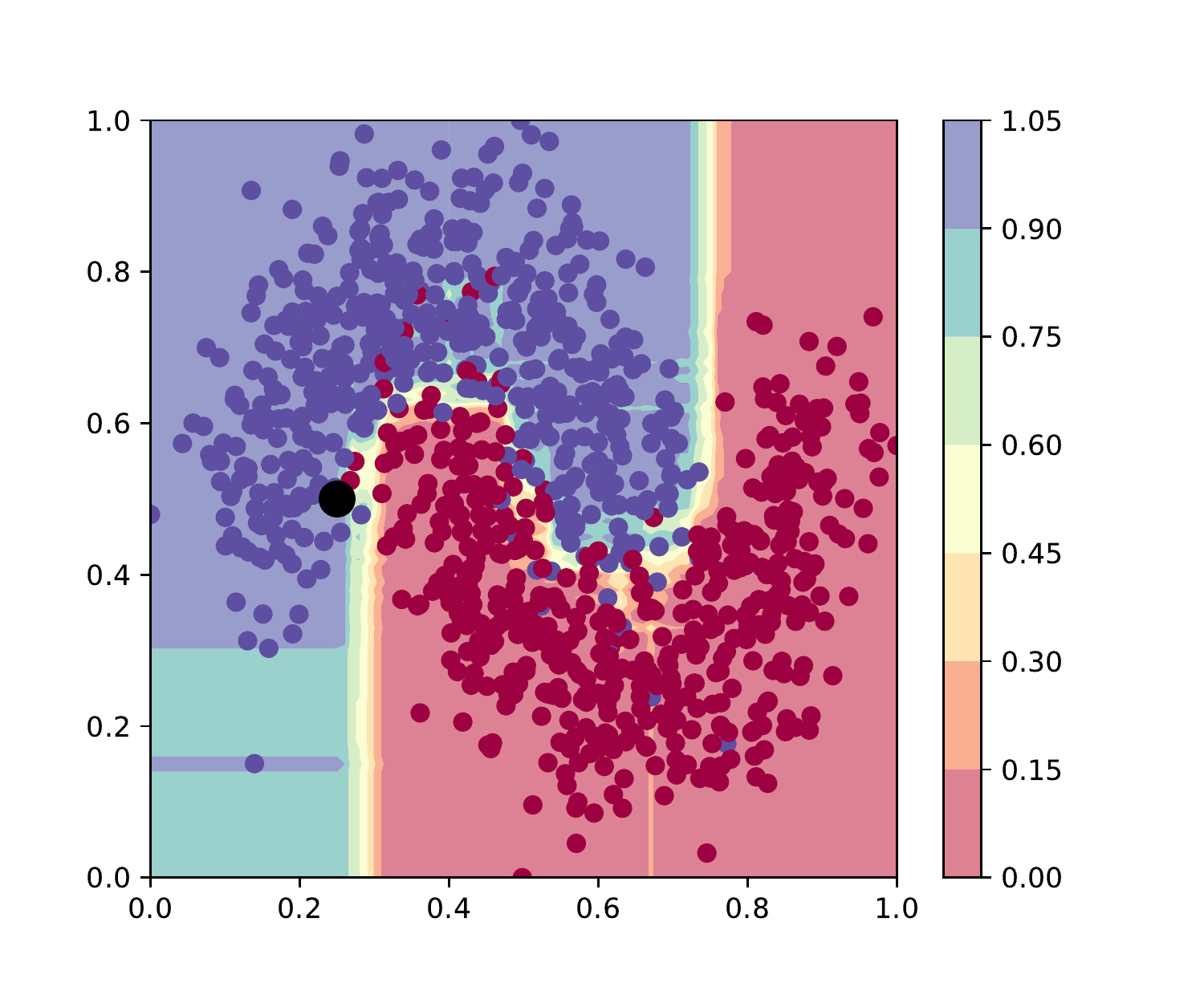}
    \caption{The two moons data set with 1000 samples. The colour of the marker indicates the ground truth label of each data point and the background shading depicts the decision boundary of the underlying black-box model (a random forest classifier). Since the model is probabilistic the colour-bar (placed to the right of the plot) provides a legend for the predicted probabilities of the blue class. The black dot represents the data point that will be explain with local surrogates (see Figures~\ref{fig:blimey_linear} and \ref{fig:blimey_tree}).\label{fig:two_moons}}
\end{figure}

To show the importance of selecting a good surrogate model and the difference in explanations that it can produce we explain a carefully selected data point from the two moons data set. %
The two moons data set -- shown in Figure~\ref{fig:two_moons} and generated with scikit-learn\footnote{\url{https://scikit-learn.org/stable/modules/generated/sklearn.datasets.make_moons.html}} -- is a synthetic 2-dimensional, binary classification data set with a complex decision boundary. %
It is suitable for this type of experiments as depending on which data point is chosen the resulting explanations can be quite diverse.%

\begin{figure}[htb]
    \centering
    \includegraphics[width=0.55\textwidth]{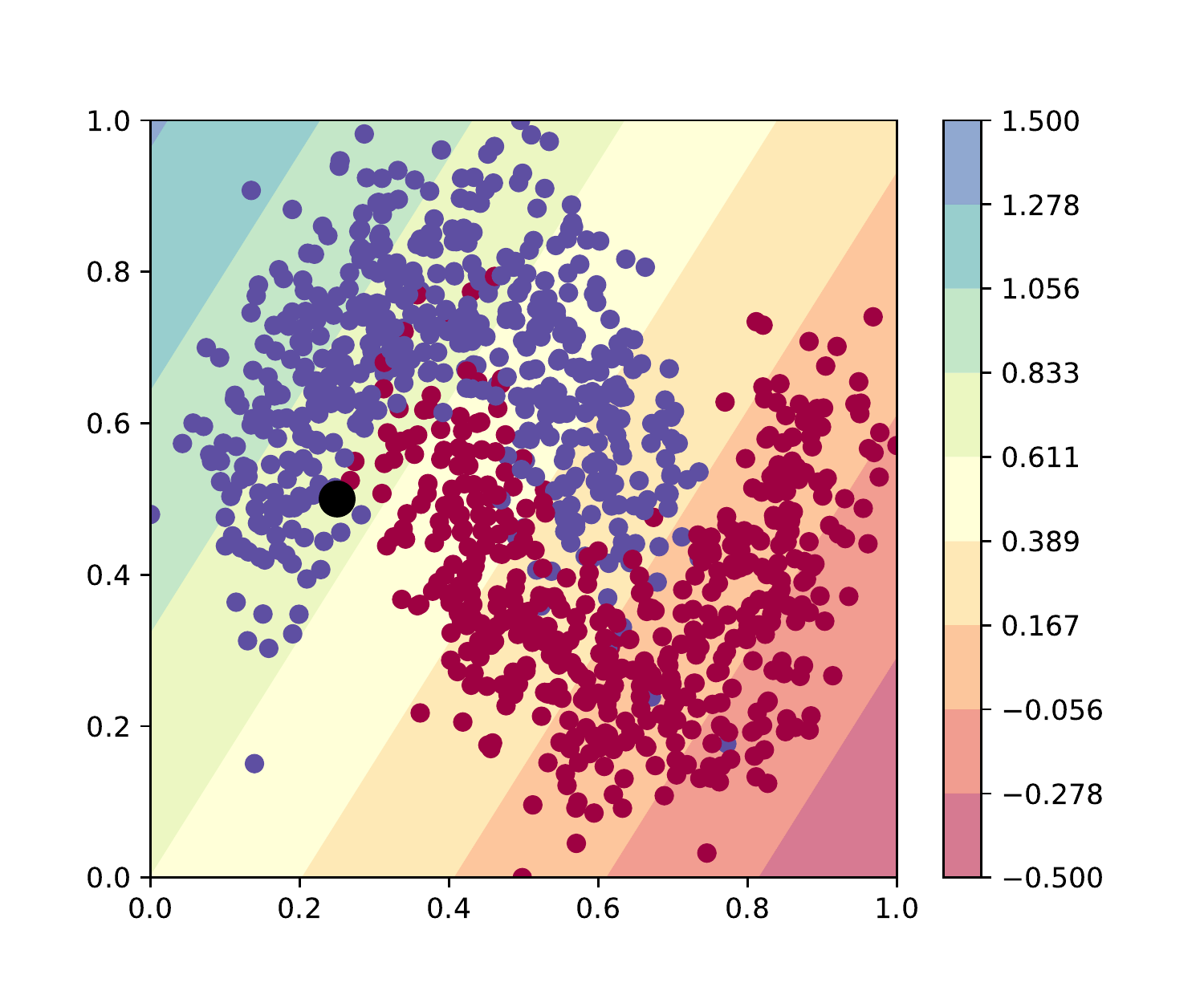}
    \caption{Linear surrogate explainer for the selected data point (black dot) -- equivalent to LIME without discretisation. The background shading represents the predicted value from the local ridge regression model (the value encoding is given in the colour-bar). The local regression model is trained to predict the probability of belonging to the blue class (outputted by the black-box model), hence the predicted values may be outside of the expected \([0, 1]\) range. If the surrogate's threshold is set at \(0.5\), the yellow bar would be partially predicted as blue, therefore incorrectly classifying the upper left part of the red cloud of points. The importance of the x-axis feature is \(-1.10\) and the importance of the y-axis feature is \(0.69\). Since an interpretable representation was not used, these values do not carry any particular meaning.\label{fig:blimey_linear}}
\end{figure}

\begin{figure}[htb]
    \centering
    \includegraphics[width=0.55\textwidth]{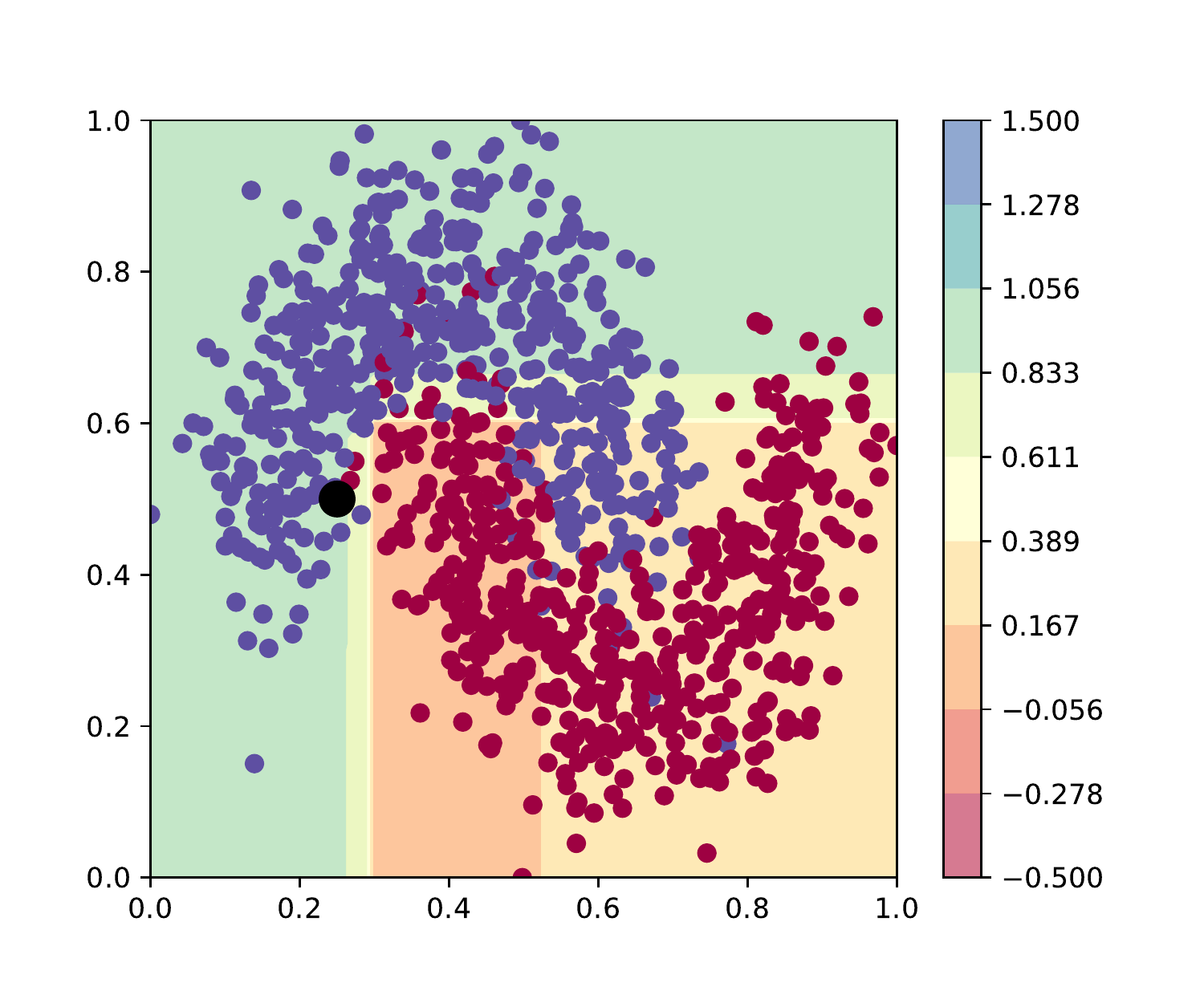}
    \caption{Decision tree-based surrogate explainer for the selected data point (black dot). The background shading represents the predicted value from the local decision tree regression model (the value encoding is given in the colour-bar). The local decision tree model is trained to predict the probability of belonging to the blue class (outputted by the black-box model). The green and light green areas have high probability of the blue class, therefore giving a good approximation of the local decision boundary, which is fairly complex for the selected data point. The orange and yellow blocks have low probability of the blue class, therefore providing a precise approximation of the red class. A possible explanation that can be derived from the local decision tree is: it is the blue class for the x-axis feature \(\leq 0.265\) or the y-axis feature \(> 0.609\); and it is the red class for the y-axis feature \(\leq 0.609\) and the x-axis feature bounded between \((0.295, 0.528]\) -- such rules can be produced for problems beyond 2-dimensions, which cannot be easily visualised.\label{fig:blimey_tree}}
\end{figure}

Figures~\ref{fig:blimey_linear} and \ref{fig:blimey_tree} show two variants of surrogate explainers -- based on a linear and a decision tree model respectively -- built for the data point marked with a black dot. %
It is clear that for complex decision boundaries a tree-based approach is superior. %
In addition to better approximating the local decision boundary (of the underlying black-box random forest classifier), a tree-based local surrogate generates locally-faithful interpretable representation from the feature splits learnt by the tree, which can be used to convey the explanation as a conjunction of logical conditions in high-dimensional spaces where visualisations are impossible.%

\end{document}